\documentclass{article}

     \PassOptionsToPackage{numbers, compress}{natbib}

     \usepackage[preprint]{neurips_2020_arxiv_2021}

\usepackage[utf8]{inputenc} 
\usepackage[T1]{fontenc}    
\usepackage{hyperref}       
\usepackage{url}            
\usepackage{booktabs}       
\usepackage{amsfonts}       
\usepackage{nicefrac}       
\usepackage{microtype}      

\newcommand{\contribsection}{\section*{Author Contribution}}
\NewEnviron{contrib}{%
  \contribsection
  \BODY
}

\usepackage{epsfig}
\usepackage{graphicx}
\usepackage{pdfpages}

\title{Paired Image-to-Image Translation Quality\\Assessment Using Multi-Method Fusion}

\author{%
  Stefan Borasinski\\
  Cyanapse Limited\\
  Brighton, United Kingdom\\
  \texttt{sborasinski@cyanapse.com} \\
  \And
  Esin Yavuz\\
  Cyanapse Limited\\
  Brighton, United Kingdom\\
  \texttt{eyavuz@cyanapse.com} \\
  \And
  S\'ebastien B\'ehuret\thanks{Corresponding author.}\\
  Cyanapse Limited\\
  Brighton, United Kingdom\\
  \texttt{sbehuret@cyanapse.com} \\
}

\begin{document}

\maketitle

\begin{abstract}
How best to evaluate synthesized images has been a longstanding problem in image-to-image translation, and to date remains largely unresolved. This paper proposes a novel approach that combines signals of image quality between paired source and transformation to predict the latter's similarity with a hypothetical ground truth. We trained a Multi-Method Fusion (MMF) model via an ensemble of gradient-boosted regressors using Image Quality Assessment (IQA) metrics to predict Deep Image Structure and Texture Similarity (DISTS), enabling models to be ranked without the need for ground truth data. Analysis revealed the task to be feature-constrained, introducing a trade-off at inference between metric computation time and prediction accuracy. The MMF model we present offers an efficient way to automate the evaluation of synthesized images, and by extension the image-to-image translation models that generated them.
\end{abstract}

\section{Introduction}

\begin{figure}
\begin{center}
\includegraphics[width=0.6\linewidth]{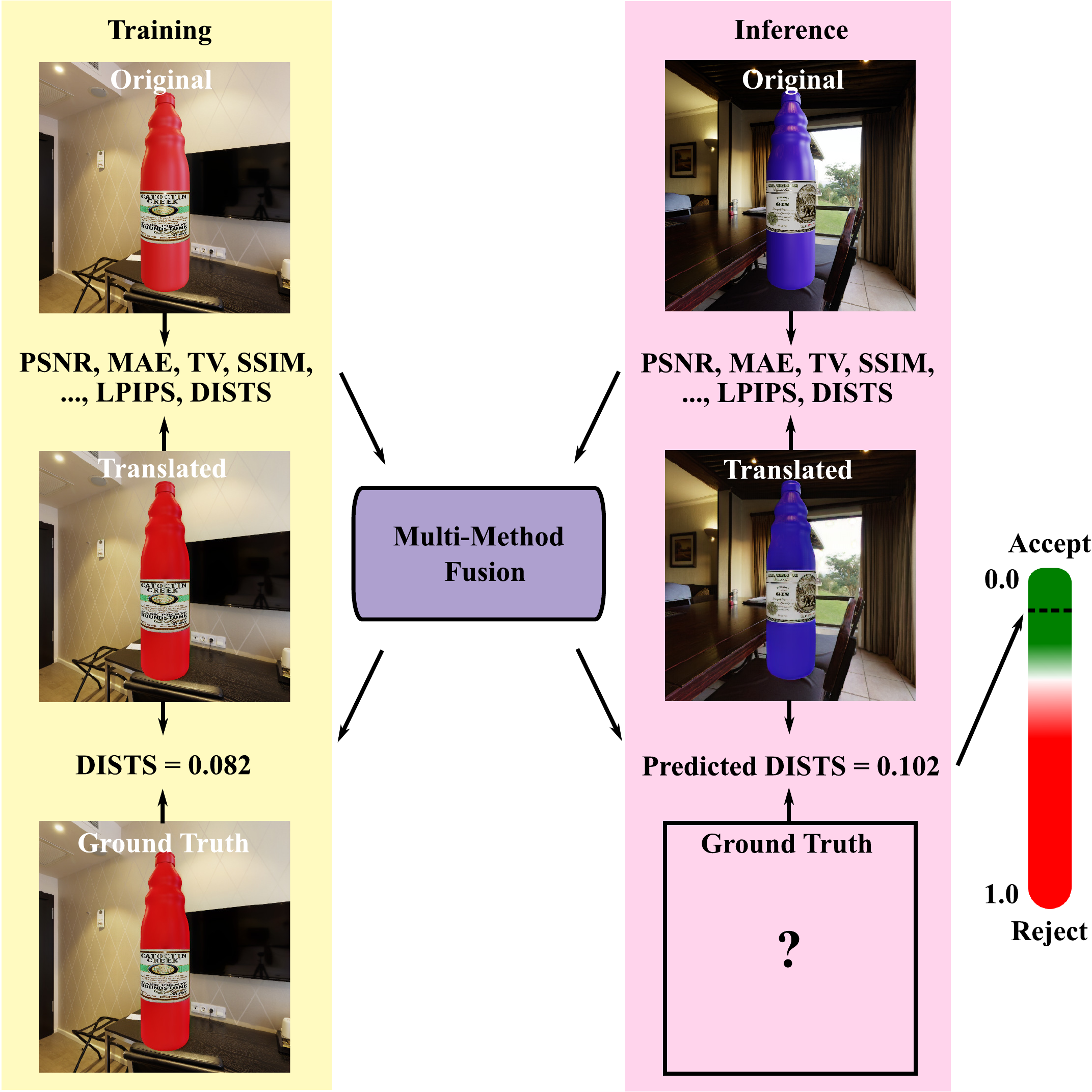}
\end{center}
   \caption{MMF process for evaluating image-to-image translation results. Perceptual differences between original and translated image pairs are evaluated using multiple IQA methods. The target metric, DISTS, is calculated between the translated image and the corresponding ground truth ({\bf Left}). The MMF model is trained on the collected IQA scores to best predict the DISTS score between an output transformation and its presumed ground truth ({\bf Middle}). At inference where ground truth images are unavailable, DISTS prediction is used as a weak indicator of transformed image quality to either accept or reject a given transformation ({\bf Right}).}
\label{fig:MMF}
\label{fig:onecol1}
\end{figure}

Image-to-image translations with style transfer enable high-level image modifications that would be considered impossible in the past \cite{gatys2016image,  isola2017image}. Evaluating the outcome, however, is not a trivial task \cite{borji2019pros}. While different sets of neural network weights can lead to high quality images on some subsets of the data domain, they may fail for others, and failure cases may be drastically bad. Efforts have been primarily focussed on post-hoc analysis, where quantitative and qualitative evaluations of model quality traditionally require a large body of generated samples. Identifying the superior model in general is not sufficient, however. The optimal model needs to be identified on an image-by-image basis at inference. But without an ad-hoc differentiator of transformed image quality, identifying the models that generated useful transformations would otherwise require manual intervention. 

Good indicators of transformed image quality should, at a minimum, be able to filter out failure cases, yielding a pool of top image candidates that could be considered reasonable transformations by humans. To this extent, Image Quality Assessment (IQA) methods can be used to assess the perceptual quality of images automatically. The motivation of an IQA-led approach to evaluating the quality of live synthesized images over other established methods lie in their ability to serve as a proxy of the human eye when discriminating at the single image level, enabling automated and continuous on-line monitoring of results. By contrast, traditional evaluation methods such as Fr\'echet Inception Distance \cite{heusel2017gans}, Average Precision, and qualitative analysis of pre-deployment results, are unsuited to assess the quality of single images in a live environment. They may indicate over a finite test set which models produce better quality transformations on average, but lack the case-by-case selectivity required when dealing with images in this context.

In scenarios where a ground truth dataset is available, for example in paired image-to-image translation during training, full-reference IQA methods offer a simple, fast, and direct means of evaluating perceptual similarity between a synthetised image (the output) and the ground truth (the target). With a view to image-to-image translation in a production setting however, where ground truth data is unavailable at the point of inference, direct comparison between the output and a corresponding target is impossible, which further precludes IQA methods from being applied in their traditional sense. Nonetheless, in many cases the original image (the source) contains aspects of image quality instructive to the output image, most notably the distribution of spatial structures. The sensitivity of IQA methods to key image quality features suggests a novel application in image-to-image translation where they can be gainfully redeployed as weak indicators of transformed image quality by comparing the source and its target only.

We describe a Multi-Method Fusion (MMF) approach to model evaluation in a similar vein to that put forward by  Liu et al. \cite{liu2012image}. Whereas Liu et al. looked to find a suitable weighting of IQA methods with the intent of predicting scores from image quality databases, we took advantage of the availability of ground truth data during training in paired image-to-image translation to propose an alternate possibility, one where IQA methods perform the role of annotator. Once a target metric of perceptual similarity has been established, the objective of MMF can simply be reformulated as predicting the target metric score between an output image and its hypothetical ground truth at inference using IQA scores extracted between source and target images. We selected DISTS \cite{ding2020image} as the target metric due to its textural sensitivities and encouraging correlation with available mean opinion score data \cite{borasinski2020iqa}. To mitigate the unpredictability of images and encompass the broad diversity of naturally occurring distortions in photos, we included a population of IQA methods to improve the likelihood that a mapping between IQA scores and DISTS will be successful. Other perceptually-aligned metrics may be more applicable depending on the specific use case.

In their MMF approach, Liu et al. \cite{liu2012image} used support vector regression, which offered superior performance over other methods at the time.  We opted for gradient boosting \cite{friedman2001greedy} using the MLJAR Automated Machine Learning (AutoML) framework instead \cite{mljar}, which offers more flexibility, and much faster computation in comparison to the scikit-learn \cite{pedregosa2011scikit} implementation of support vector regression, owing to the use of GPUs.

\section{Related Work}
\subsection{Image Quality Assessment}

IQA was originally envisioned as a means to objectively evaluate broadcast images with reference to the human visual system \cite{goldmark1940quality, horton1929electrical, jesty1953television}. The field of IQA has since seen a multitude of novel applications \cite{wang2011applications}, and even more so during the last decade as artificial intelligence methods for image generation and manipulation have been shown to produce plausible results. In deep learning, loss functions motivated by perceptual considerations neatly illustrate the tight confluence between IQA and image processing systems.  Image denoising and restoration tasks have been successfully addressed by directly optimizing IQA methods \cite{ding2021comparison, zhao2016loss}, and conversely new IQA methods themselves have been adapted from losses that incorporate deep feature representations \cite{ding2020image, zhang2018unreasonable}. 

In order to act as a proxy for human judgments of image similarity, IQA algorithms aim to detect a certain type of difference between a transformed image and a reference image. A detailed taxonomy of the most widely known IQA methods can  be found in \cite{ding2021comparison}. Pixel-based error visibility methods such as mean absolute error (MAE), peak signal-to-noise ratio (PSNR), total variation (TV), normalized Laplacian pyramid distance (NLPD) \cite{laparra2016perceptual}, and most apparent distortion (MAD) \cite{larson2010most} each detect a certain type of distance between pixels. These methods are the most straightforward, but they are also the least correlated with human perception. Structural similarity-based methods are variants of the original structural similarity (SSIM)\cite{wang2004image} algorithm. Structural similarity is inspired by the human visual system, and it has been the most popular choice of metric in IQA.  Structural similarity-based methods include SSIM, multi-scale structural similarity (MS-SSIM) \cite{wang2003multiscale}, feature similarity (FSIM) \cite{zhang2011fsim},  gradient magnitude similarity deviation (GMSD) \cite{xue2013gradient}, and visual saliency induced quality index (VSI) \cite{zhang2014vsi}. Information-theoretic methods such as visual information fidelity (VIF) \cite{sheikh2006image} and spatial domain VIF (VIFs) are based on mutual information. Learning-based methods are based on deep neural network representations of images. Benefiting from the recent advances in the field, these methods now provide a powerful alternative to SSIM-based methods. Learning-based methods include deep image structure and texture similarity (DISTS) \cite{ding2020image2}, and learned perceptual image patch similarity (LPIPS) \cite{zhang2018unreasonable}.

\subsection{Multi-Method Fusion}

MMF is in essence a classical machine learning task whereby the features are the scores from full-reference image quality metrics.  Fusion-based methods are based on a super-evaluator where fused individual image quality scores are often matched with human opinion scores \cite{liu2012image, ma2019blind}. Collecting human opinion scores is,  however,  a time-consuming and expensive task. Training human assessors is often not trivial as human judgment is affected by many different factors and is highly variable.  Where reference images are available,  a more reliable outcome can be achieved by using matching ground truth images.

\subsection{Gradient Boosting}

Gradient-boosted decision trees \cite{friedman2001greedy} allow fast and efficient prediction for tabular data through ensembling of multiple weak models to solve supervised learning problems. Unlike other modes of regression, decision tree-based algorithms are able to natively handle the multi-collinear nature of features calculated from the same pair of images, eliminating the need for further pre-processing. Recent variations include CatBoost \cite{dorogush2018catboost}, LightGBM \cite{ke2017lightgbm} and XGBoost \cite{chen2016xgboost} to improve speed and accuracy, each using a different strategy for ensembling. We combined these three methods to diversify the learning process.

\subsection{Automated Machine Learning}

In the recent years, AutoML has become a popular approach to perform feature engineering, design model archtectures and optimize hyperparameters with state-of-the-art algorithms. To address the inherently high-dimensional feature and parametric spaces of our MMF implementation and to facilitate the model development cycle, we opted for the use of an AutoML framework. Among the popular AutoML frameworks, MLJAR offers a flexibile training pipeline capable of achieving high performances on various datasets. It combines K-Means centers and Golden Features search with advanced feature selection to enrich datasets, Hill-Climbing to fine-tune final models, and Greedy Search over base models to compute ensembles. We therefore used the \textsc{mljar-supervised} \cite{mljar} Python package to build our MMF model.

\section{Proposed Method}

Image-to-image translation involves mapping an image in $ \mathcal{X}$ into domain $ \mathcal{Y}$ by a mapping function $F : \mathcal{X} \rightarrow  \mathcal{Y}$. In paired image-to-image translation, a ground truth image $y \in \mathcal{Y}$ is available for each source image $x \in \mathcal{X}$ at training time. The perceptutal distance function $E = D(F(x),y)$ indicates the level of similarity between translated images and their ground truth targets. Our goal is to train a MMF model to estimate a perceptual distance function $\hat{E} = D(F(\hat{x}),\hat{y})$ and further assess the success of tranformation by $F$ for a novel image $\hat{x}$ and its hypothetical and unavailable ground truth $\hat{y}$.

Our approach is illustrated in Figure \ref{fig:MMF}. We propose a MMF strategy that combines various IQA metrics that assess the similarity between source images $x_{n}$ and their translated outputs $F(x_{n})$. The similarity between $F(x_{n})$ and corresponding ground truth targets $y_{n}$ is estimated by calculating the value of the perceptual distance function $E_{n}$, as quantified by the DISTS score (Fig. \ref{fig:MMF}, Left). Lower DISTS scores indicate more similarity between the translated output and the ground truth images. IQA methods give different scores for each source image $x$ as a result of the nature of the differences between $x$,$y$ and $F(x)$. The trained MMF model leverages these differences, as reflected by IQA scores, by combining them and finding optimal weights using a supervised learning strategy (Fig. \ref{fig:MMF}, Middle) to estimate $\hat{E}$ between a novel source image $\hat{x}$ and its presumed ground truth $\hat{y}$ during inference (Fig. \ref{fig:MMF}, Right). As in \cite{liu2012image}, the rationale behind MMF in image-to-image translation is to weight a collection of complementary yet diverse signifiers of transformed image quality into a singular, more powerful evaluator.

\section{Experiments}

Our MMF process collects and integrates an ensemble of up to 14 IQA methods, which were used with their default trained hyperparameters where applicable. Toward this, 12 metrics were adapted or directly imported without further modification from the \textsc{scikit-image} \cite{scikit-image} and \textsc{IQA-pytorch} \cite{ding2021comparison} Python packages. This includes PSNR, SSIM, MS-SSIM, FSIM, VSI, GMSD, NLPD, MAD, VIF, VIFs, LPIPS and DISTS metrics. The last two metrics, TV and MAE, were reimplemented in NumPy \cite{harris2020array}. TV was reimplemented as a full reference metric, converted into a ratio (TV Ratio) by dividing the sum total variation in the input image by that in the output transformation. MAE was calculated as the mean of the absolute difference between individual pixel values of two images. MAD was not included in the MMF model for all tasks due to its time complexity.

As the number of features is restricted by the total time to compute IQA metrics at inference, one immediate way to improve predictive accuracy of the regression algorithm is to augment the dataset by computing K-Means centers and including linear combinations of original features, dubbed Golden Features \cite{slezak1999classification}. To further sharpen predictive accuracy, the MMF model not only fuses the aforementionned IQA methods themselves, but also ensembles weak evaluators. To this end, a family of gradient tree boosting algorithms, LightGBM, CatBoost and XGBoost, were entered into the training cycle. We used the MLJAR AutoML framework in the 'compete' mode to augment the data, train and combine the models under the regimen described below.

To enrich the dataset, K-Means centers were computed with optional scaling as needed. The information about distance to K-Means centers and center number was added to each sample and the best performing models were trainined with this data. For Golden Features, original feature pairs were either subtracted, added, multiplied or turned into ratios to create new features. To select the most useful linear combinations of features from all possible unique pairs that are capable of improving predictions, all possible pairs were examined. Independent decision trees were trained on the task with a maximum depth of 3 on a single combination of features at a time using a subsample of 5,000 points split equally between the train and test sets. Test mean squared error was calculated for each new feature and the top features were entered into the primary dataset in order to retrain the best performing models.

The framework's feature selection algorithm works as follows. Random features were inserted into the data and the permutation-based features importance was computed. The features with importance less than a random variable were removed. The ML models are trained using selected features only.

Multiple models were trained for each of the LightGBM, CatBoost and XGBoost algorithms. For each algorithm, a single base model was trained using default hyperparameter settings, and multiple additional models were trained using randomly sampled hyperparameters. Validation was performed automatically using the most adapted algorithm with the root-mean-square error metric, including 5 to 10-fold cross-validation and 80/20 train/test split hold-out validation. To fine-tune models, the framework performed a local search over hyperparameters via hill-climb. The framework additionally trained stacked models using out-of-fold predictions to extend training data, and boost-on-errors models where sample weights were boosted on the errors from the best models. The greedy search algorithm outlined in \cite{caruana2004ensemble} combined the best performing models, consisting of stacked, boost-on-errors and ensembled LightGBM, CatBoost and XGBoost models, some of which incorporated K-Means centers or Golden Features. To reduce overfitting, the ensembling algorithm initialized non-empty sorted ensembles, selected base models with replacement, and further bagged base models during selection. It is the final stacked ensemble produced by the framework that constitute the MMF model presented in this paper.

We demonstrate the results on two image-to-image translation tasks.  The first task is base color extraction from synthetic images of consumer goods via style transfer using a variant of Pix2Pix \cite{isola2017image}. The synthetic dataset was by provided by ZEG.ai\footnote {ZEG.ai Ltd. (London, United Kingdom).} (commercially restricted) and includes pairs of synthetic bottle images under complex and albedo-like diffuse-only lighting conditions. For this dataset we show the error visibility maps, and MMF evaluation. The second task involves day to night transformations using a pre-trained Pix2Pix network\cite{isola2017image}. We used the dataset \cite{laffont2014transient} and trained model weights publicly available at \url{https://github.com/junyanz/pytorch-CycleGAN-and-pix2pix}. For this dataset we show MMF evaluation and samples of transformations with the predicted scores in the best, average and worst DISTS score ranges.

All the analysis and benchmarks are performed on raw neural network outputs, which correspond to 256x256 pixels resolution RGB images.

\subsection{Error Visibility Maps}

\begin{figure*}
\begin{center}
   \includegraphics[width=0.9\linewidth]{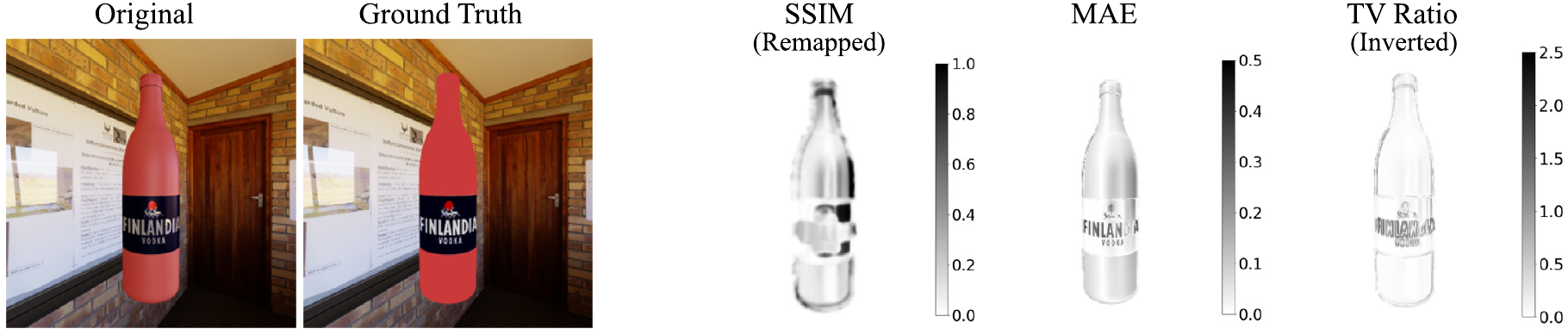}
\end{center}
   \caption{An example original and ground truth image pair from the test dataset for the base color extraction task {\bf (Left)}, and corresponding metric error visibility maps for SSIM, MAE and TV Ratio {\bf (Right)}. For illustrative and comparative purposes, SSIM error signals have been remapped to $1 - SSIM$ and clamped between [0, 1], and TV Ratio has been inverted. While SSIM operates on grayscale images, MAE and TV Ratio operate on RGB images. Images were therefore converted into grayscale to compare the magnitude of pixelwise error as indicated by gray intensity, giving scores of $SSIM = 0.970$, $MAE = 0.00867$ and $TV Ratio = 1.0146$. MAE and TV Ratio were also caclulated on RGB images, giving scores of $MAE_{RGB} = 2.649$ and $TV Ratio_{RGB} = 1.020$. Lower remapped SSIM scores, MAE scores and inverted TV Ratio scores suggest better transformations.}
\label{fig:maps}
\end{figure*}

The differences between the types of errors picked up by IQA methods can be seen in the  error visibility maps that represent the per-pixel metric scores. Figure \ref{fig:maps} shows an example original and ground truth image pair (Fig. \ref{fig:maps}, Left) and corresponding error visibility maps (Fig. \ref{fig:maps}, Right) for three different IQA methods (SSIM, MAE, TV Ratio) for the base color extraction task. SSIM, which is sensitive to changes in luminance and structure, penalizes the removal of lighting from the label as well as loss in detail caused by flattening the bottom edge of the cap. Compared to the other methods, error signals extracted by SSIM are less finely localized due to local patch averaging, producing blurry error outlines. The MAE map primarily captures changes in chromaticity, and does not provide information on the structure of an image. By contrast, TV captures absolute differences in gradient magnitude across neighboring pixels and informs on the local structure of an image, highlighting areas that have become smoother, and leading to an accentuation of borders. A decrease in the output TV is expected for successful transformations for this task due to lighting smoothing in the output transformation, further leading to an increase of the TV Ratio.

\subsection{IQA Metrics Benchmarks}

\setlength{\tabcolsep}{0.1em} 
\begin{table}
\small
\begin{center}
\label{table:benchmark}
\begin{tabular}{|l|c|c|c|c|}
\hline
Method & Wall-Clock Time (ms) & Original vs. Ground Truth & Original vs. Translated & Translated vs. Ground Truth\\
\hline\hline
PSNR $\uparrow$ & 0.315 $\pm$ 0.018 & 24.474 $\pm$ 4.127  & 27.918 $\pm$ 4.569 & 23.724 $\pm$ 3.793\\
TV Ratio $\uparrow$ & 0.850 $\pm$ 0.026 & 1.071$\pm$ 0.030  & 0.960 $\pm$ 0.050 & 0.896 $\pm$ 0.052\\
MAE $\downarrow$ & 0.448 $\pm$ 0.065 & 16.504$\pm$ 7.375 & 120.831 $\pm$ 48.809 & 122.223 $\pm$ 45.918\\
SSIM $\uparrow$ & 1.018 $\pm$ 0.094 & 0.943 $\pm$ 0.024 & 0.955 $\pm$ 0.040 & 0.917 $\pm$ 0.043\\
MS-SSIM $\uparrow$ & 3.597 $\pm$ 0.128 & 0.866 $\pm$ 0.124  & 0.948 $\pm$ 0.076 & 0.873 $\pm$ 0.119\\
FSIM $\uparrow$ & 121.504 $\pm$ 3.215 & 0.945 $\pm$ 0.026  & 0.970 $\pm$ 0.020 & 0.932 $\pm$ 0.026\\
VSI $\uparrow$ & 15.862 $\pm$ 0.633 & 0.966 $\pm$ 0.019  & 0.990 $\pm$ 0.008 & 0.966 $\pm$ 0.017\\
GMSD $\downarrow$ & 0.989 $\pm$ 0.038 & 0.130 $\pm$ 0.036  & 0.049 $\pm$ 0.027 & 0.124 $\pm$ 0.033\\
NLPD $\downarrow$ & 5.384 $\pm$ 0.110 & 0.304 $\pm$ 0.121  & 0.198 $\pm$ 0.092 & 0.319 $\pm$ 0.117\\
MAD $\downarrow$ & 179.336 $\pm$ 3.582 & 110.802 $\pm$ 21.870  & 83.518 $\pm$ 25.858 & 127.658 $\pm$ 19.980\\
VIF $\uparrow$ & 74.309 $\pm$ 2.624 &  0.780 $\pm$ 0.060  & 0.676 $\pm$ 0.106 & 0.585 $\pm$ 0.085\\
VIFs $\uparrow$ & 6.159 $\pm$ 0.134 & 0.820 $\pm$ 0.048  & 0.689 $\pm$ 0.102 & 0.598 $\pm$ 0.081\\
LPIPS $\downarrow$ & 28.924 $\pm$ 0.864 & 0.051$\pm$ 0.017  &0.062 $\pm$ 0.046 & 0.100 $\pm$ 0.046\\
DISTS $\downarrow$ & 28.860 $\pm$ 0.751 & 0.095 $\pm$ 0.026  & 0.062 $\pm$ 0.032 & 0.117 $\pm$ 0.030\\
\hline
\end{tabular}
\bigskip
\caption{Average metric scores $\pm$ standard deviation (n=50) and computation clock time for the base color extraction task. The vertical arrows represent the direction of variation of each metric score, with upward arrows indicating increased similarity with larger values, and downward arrows indicating increased errors with larger values. Original vs. Ground Truth scores indicate the zone of viability for a successful transformation. Original vs. Translated scores show how close on average transformations approach that zone. Translated vs. Ground Truth scores indicate the average similarity between transformation and target.}
\end{center}
\end{table}

In order to evaluate the performance of each IQA metric function, we ran several benchmarks. Individual metric scores provide a baseline to assess whether the calculated score for a given sample pair is in a reasonable range, which can in turn be used as an additional factor to predict the transformation quality. It is the role of the trained MMF model to weight the information provided by these metrics to appropriately predict similarity between an output image and its hypothetical ground truth.

For the base color extraction task, results are shown in Table \ref{table:benchmark}. Each metric was calculated 50 times over a total of 707 unique combinations of original, translated and ground truth images. Calculation times were then averaged for each metric. The 14 IQA metrics could be calculated in a combined wall-clock time of less than a third of a second per image, with the option of further reductions by tweaking their parameters if necessary.

For the day to night translation task, results are shown in Table \ref{table:benchmark2}. Likewise, metrics were calculated a total of 50 times between the original, translated and ground truth images, then averaged. The MAD metric was excluded owing to expense of calculation.

For metrics that indicate increased similarity with higher values, as represented by upward arrows (and vice versa for metrics that indicate decreased similarity with higher values, as represented by downward arrows, such as DISTS), high metric scores between an original image and translation can be a result of very weak changes, which generally indicate an unsuccessful translation. By contrast, low metric scores suggest that the structure or the texture in the translated image is drastically different compared to that of the original image.

\begin{table}
\small
\begin{center}
\begin{tabular}{|l|c|c|c|}
\hline
Method & Original vs. Ground Truth & Original vs. Translated & Translated vs. Ground Truth\\
\hline\hline
PSNR $\uparrow$ & 9.169 $\pm$ 2.380  & 9.451 $\pm$ 1.879 & 13.668 $\pm$ 3.563\\
TV Ratio $\uparrow$ & 1.038 $\pm$ 0.221 & 0.988 $\pm$ 0.090  & 1.047 $\pm$ 0.139\\
MAE $\downarrow$ & 103.997 $\pm$ 17.072 & 98.121 $\pm$ 15.766 & 113.335 $\pm$ 38.449\\
SSIM $\uparrow$ & 0.365 $\pm$ 0.126 & 0.391 $\pm$ 0.071 & 0.435 $\pm$ 0.120\\
MS-SSIM $\uparrow$ & 0.264 $\pm$ 0.140  & 0.295 $\pm$ 0.104 & 0.309 $\pm$ 0.170\\
FSIM $\uparrow$  & 0.641 $\pm$ 0.056  & 0.646 $\pm$ 0.053 & 0.695 $\pm$ 0.073\\
VSI $\uparrow$ & 0.854 $\pm$ 0.032  & 0.865 $\pm$ 0.028 & 0.877 $\pm$ 0.041\\
GMSD $\downarrow$ & 0.264 $\pm$ 0.032  & 0.263 $\pm$ 0.038 & 0.245 $\pm$ 0.045\\
NLPD $\downarrow$ & 0.923 $\pm$ 0.120  & 0.871 $\pm$ 0.163 & 0.803 $\pm$ 0.207\\
MAD $\downarrow$ & N/A  & N/A & N/A\\
VIF $\uparrow$ &  0.056 $\pm$ 0.031  & 0.043 $\pm$ 0.013 & 0.044 $\pm$ 0.019\\
VIFs $\uparrow$ & 0.097 $\pm$ 0.047  & 0.094 $\pm$ 0.034 & 0.051 $\pm$ 0.028\\
LPIPS $\downarrow$ & 0.595$\pm$ 0.083  &0.672 $\pm$ 0.045 & 0.644 $\pm$ 0.054\\
DISTS $\downarrow$ & 0.325 $\pm$ 0.073  & 0.404 $\pm$ 0.050 & 0.362 $\pm$ 0.051\\
\hline
\end{tabular}
\bigskip
\caption{Average metric scores $\pm$ standard deviation (n=50) for the day to night translation task. MAD was omitted due to time complexity. For vertical arrows, see caption of Table \ref{table:benchmark}.}
\label{table:benchmark2}
\end{center}
\end{table}

\subsection{MMF Model Evaluation}

\begin{figure}
\begin{center}
   \includegraphics[width=0.5\linewidth]{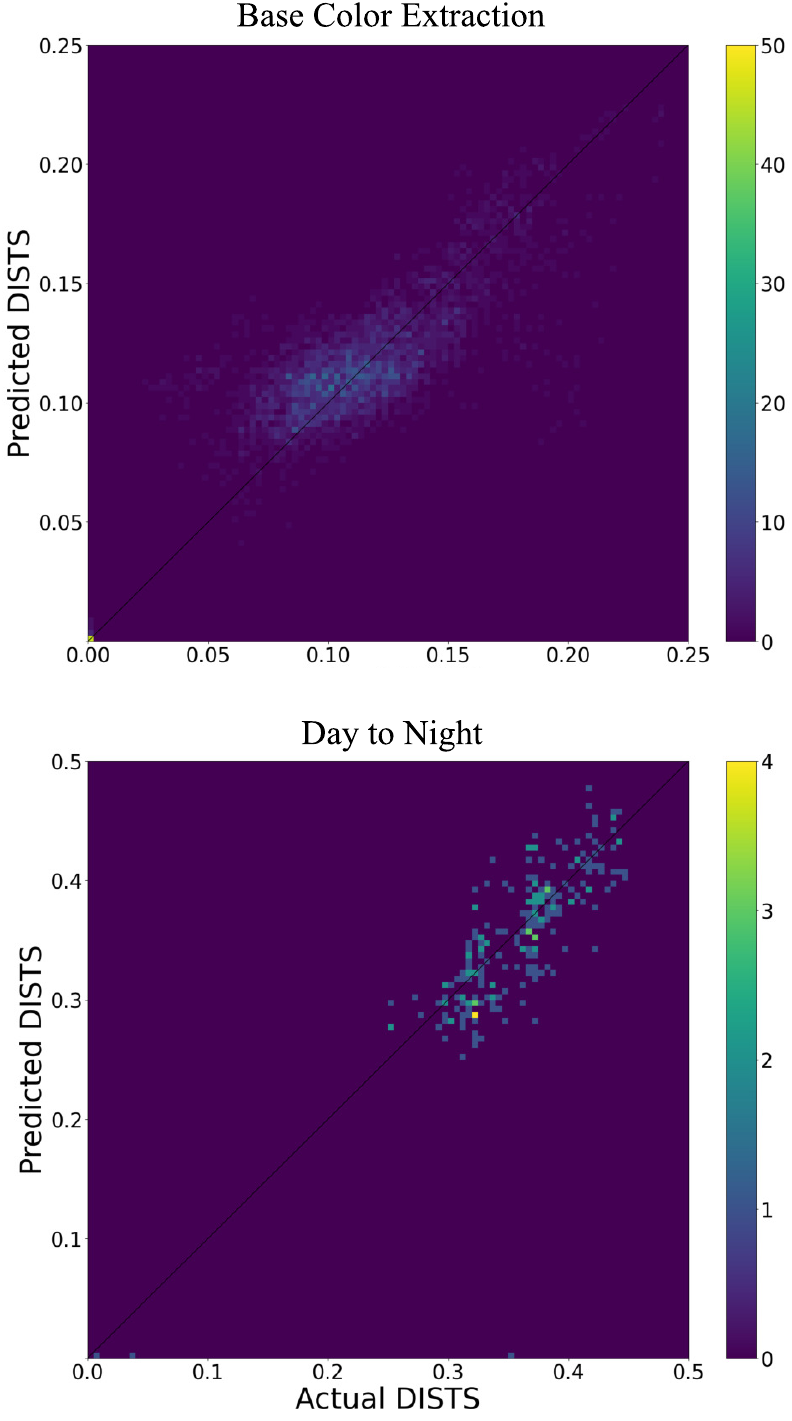}
\end{center}
   \caption{Correlation between actual and predicted DISTS. The 100x100-bin heatmaps shows a visual representation of the points contributing to $r^{2}$ score in relation to the y=x line of best fit. For the base color extraction task, $r^{2}=0.666$ ({\bf Top}), and for the day to night translation task, $r^{2}=0.722$ ({\bf Bottom}).}
\label{fig:corr}
\end{figure}

The MMF model consisted of a trio of gradient-boosted regression algorithms, LightGBM, CatBoost and XGBoost, that were trained and ensembled on image quality scores between the input and the output, to predict similarity quantified by DISTS between the output and the ground truth. Correlation between predicted vs. actual DISTS is shown in Figure \ref{fig:corr} for the base color extraction and day to night translation tasks.

For the base color extraction task (Fig. \ref{fig:corr}, Top), the complete training dataset for the MMF model was created by applying each of the 13 metrics (all metrics except MAD) to 31,108 input and output transformation pairs generated from 44 image-to-image translation models over a total of 707 unique input images, and 707 input and ground truth pairs, for a total of 31,815 image pairs.  The inclusion of input and ground truth pairs ensured that the model is trained on metrics scores calculated from perfect transformations. DISTS was then calculated for each output transformation and ground truth pair. From the 707 source images, 10\% were split into a test set and the remainder used for 5-fold cross-validation. Since the sole alteration in the task is textural, similarity between the input and ground truth images remained high in general. The bulk of test instances recorded an actual DISTS score of between 0.075 to 0.250, with variation within that range in both directions contributing to the central mass of the heatmap. A handful of instances at the lower and upper end of actual DISTS deviate most notably from the line of best fit, attributable to a tendency to fit scores within this aforementioned median range. On average, transformations scored a DISTS value of 0.117 with a standard deviation of 0.030. The ensembled MMF model was able to predict DISTS to a mean absolute error of 0.0153, resulting in an $r^{2}$ score of 0.666. Transformations that scored lower showed a general qualitative superiority over those that scored higher.

For the day to night translation task (Fig. \ref{fig:corr}, Bottom), the complete training dataset comprised of 2,287 input and output transformation pairs, and 51 input and ground truth pairs, for a total of 2,338 image pairs. The input and ground truth images were added to match the ground truth pairing ratio of 1/45 from the base color extraction training dataset. Unlike in the previous task, each source image has multiple feasible ground truth transformations, meaning that DISTS score varies for the same source image across multiple pairs. Nevertheless, target DISTS scores could be successfully predicted and used for ranking the resulting transformations. Compared to a possible ground truth, the average day to night transformation scored a DISTS similarity of 0.362 with a standard deviation of 0.051. DISTS scores were predicted with a mean absolute error of 0.0250, resulting in an $r^{2}$ score of 0.722.

\begin{figure*}
\begin{center}
   \includegraphics[width=0.8\linewidth]{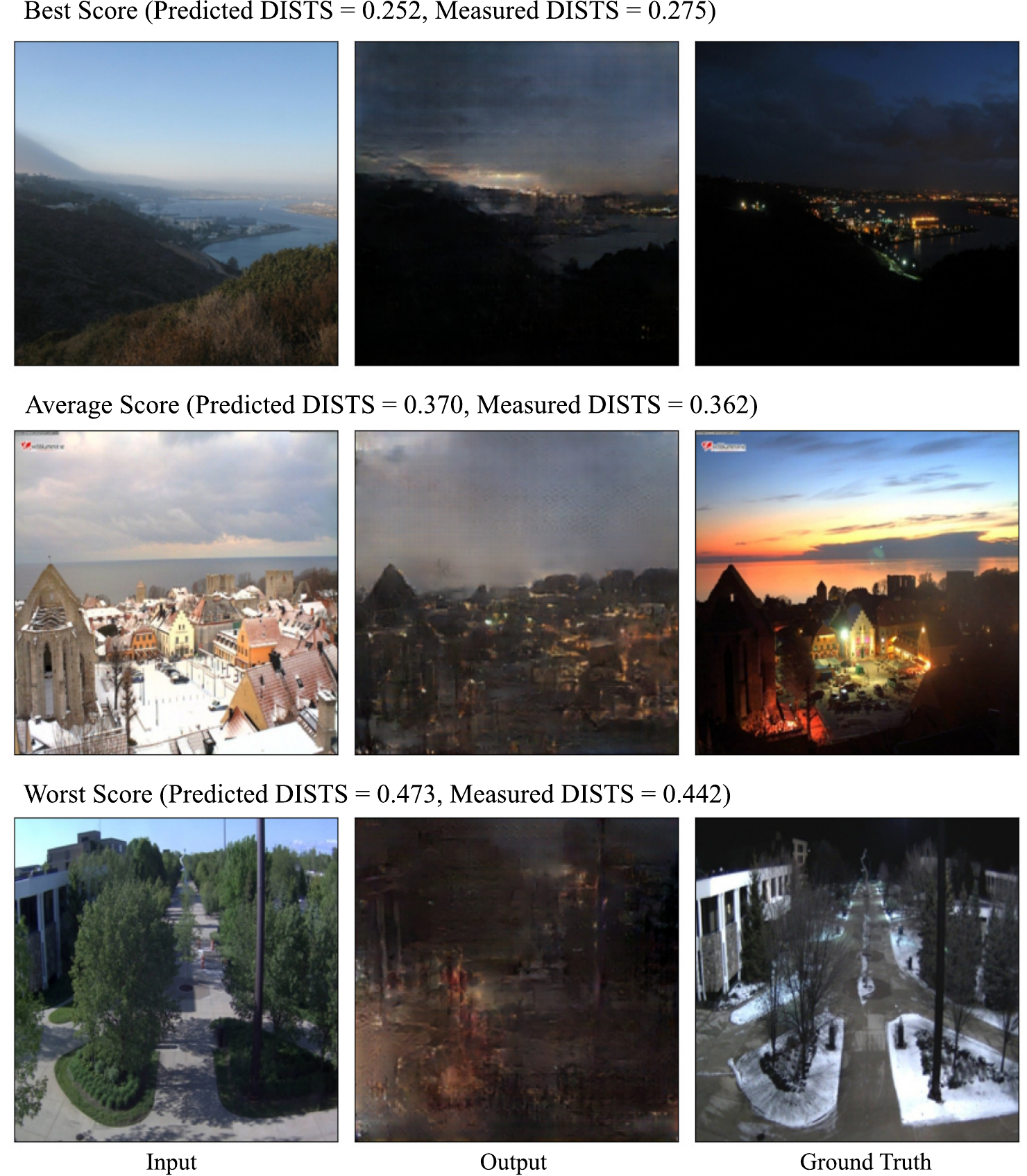}
\end{center}
   \caption{Samples with best (lowest, {\bf Top}), average ({\bf Middle}) and worst (highest, {\bf Bottom}) DISTS scores for the day to night translation task.}
\label{fig:day2night}
\end{figure*}

Image translations with best, average and worst examples of predicted DISTS are shown in Figure \ref{fig:day2night}. Based on the MMF results, day to night translations tend to work better for certain source images than others (Fig. \ref{fig:day2night}, Top), which is possibly related to content complexity and the fact that the day to night model favors safe translations where fine details are not easily perceptually detectable. Images in the worst DISTS band (Fig. \ref{fig:day2night}, Bottom) are often heavily distorted and can be discarded off-hand. Outcomes scoring around average for the dataset show some degree of distortion (Fig. \ref{fig:day2night}, Middle), and likely require further investigation before being disregarded. Predicted and actual DISTS scores for five samples in the best, average and worst score bands are included in Table S1. Corresponding images are shown in Figures S1, S2 and S3, respectively.

\section{Discussion}

In an extension to the MMF solution adopted by Liu et al. \cite{liu2012image}, who applied support vector regression to ensemble IQA methods, we found that a weighted combination of gradient-boosted regression algorithms produced the best predictive accuracy. In training the MMF model, the task appeared to be feature constrained, as removing a subset of IQA methods increased the mean absolute error. Metric calculation time thus represents the major limiting factor for MMF in a live system. 

Whilst the MMF model cannot address all aspects of transformed image quality, combined IQA methods can indicate whether a transformation falls within the zone of viability for it to be considered successful.  IQA methods that are used in this study do not comment on whether specific facets of the transformation are accurate, for example whether colors have been transformed correctly. Per-pixel errors can be detected by AI-based methods where it is possible to investigate and visualize individual filters, or combine filters to achieve a single score. Better understanding of this type of errors needs more in-depth analysis which is beyond the scope of this work. In the absence of a fitting ad-hoc method of quantitative evaluation in image-to-image translation, MMF represents a reasonable and highly practical solution where paired datasets are available during training.

The choice of target metric, DISTS, was informed by two key factors. In the base color extraction task, the image-to-image mapping repaints the focal object in its base colors, demanding a consistency between textures in output and ground truth images. DISTS is the first metric of its kind to explicitly accommodate textural regularities in its calculation, and as such offers clear utility in assessing the extent to which they appear similar. DISTS sports competitive performance on standard IQA databases and offers a robustness to mild geometric distortions, accrediting it specialist status in evaluating GAN-generated images. Furthermore, in a separate study of transformation outcomes, we observed that it was the metric maximally correlated with mean opinion score  \cite{borasinski2020iqa}.

There are a couple of considerations when interpreting the MMF results. Where a lower DISTS score would normally suggest a better transformation, it may also indicate that only a slight transformation has taken place and the output image is close to both the input and ground truth images. In this situation, IQA error signals between the input and output images are expected to be low. On the other hand, a higher DISTS score may indicate the presence of comprising distortions to transformed image quality, which often indicate deviations from the image domain or presence of heavy artifacts. Automatically filtering out failure cases via DISTS thresholding represents an intuitive way to leverage the MMF model in many use cases.

Whilst this paper presented DISTS as the target of MMF, for extended application the choice of which metric to predict should be adapted to the idiosyncrasies of the mapping at hand. Domain-relevant knowledge regarding the prerequisite features of a successful transformation may help inform which metrics can act as useful differentiators of perceptual quality. Likewise, collecting limited human evaluation data in order to calculate correlation coefficients for a range of metrics can assist in this decision. In some cases, it may be viable to train multiple MMF models to predict complementary metrics. For example, one target metric may be more sensitive to the semantic content of the transformed image, i.e. using a measure of pixelwise error to assess whether color has been appropriately transposed, where another metric such as SSIM is then concerned with the integrity of spatial structures.

The effective contribution of linear recombinations of select features suggests that the MMF model may be primarily constrained by the number of features available for the task. In turn, the limiting factor on available features is their sum computational time. With this, users need to decide whether they are more interested in model speed or model accuracy. If two further time-intensive methods are removed from calculation, FSIM and VIF, reducing sum computation time to less than a tenth of a second per image, mean absolute error increases from 0.0153 to 0.0156 and $r^{2}$ drops from 0.666 to 0.641. This does also mean however that the reverse is true, where the inclusion, at a timecost, of new methods that have become available stands to improve the accuracy of future MMF models.

As a technique, many of MMF's constituent methods inform about the statistical properties of an image pair, whether structures remain intact, if textures are consistent between images, whether there are global changes in luminance, and so on. Dependent on the methods chosen, MMF can only offer limited comment, if any, on a transformation's wider semantics, for example when inferring a bottle's true color for an albedo transformation. For many metrics, the difference between correct and incorrect shades may be marginal to non-existent in their given score. For this reason, metrics such as mean absolute error, which promote pixel values closer to the original, are valuable in this semantic. In more extreme transformations, such as in the Pix2Pix day to night model, there is no metric which can evaluate whether a source of light is appropriately positioned or not within the wider context of an image. For this reason, it is suggested that MMF be used as a filtering tool to reject outright those transformations that are unviable in any sense due to excessive distortion to better direct attention resources to those with a reasonable chance of success.

\section{Conclusion}

We presented a multi-method fusion approach to assess the quality of transformations that are generated by image-to-image translation models where reference ground truth data is available during training. We demonstrated that the MMF model, which was trained on a collection of IQA metrics between input and transformed images, can successfully predict the target metric score (in our case, DISTS) between a transformed image and its presumed ground truth. This score can then be used to predict the quality of the output during inference where ground truth data is unavailable.

\begin{contrib}
Stefan Borasinski conducted the research, performed numerical experiments, and contributed to the initial version of the manuscript. Esin Yavuz designed the study, supervised the research, cured the data for numerical experiments, and wrote the manuscript. S\'ebastien B\'ehuret supervised the study and research, set up the infrastructure for numerical experiments, and wrote and revised the manuscript.
\end{contrib}

\begin{ack}
This work was supported by the UK's innovation agency (Innovate UK Smart Grants February 2019, Project Number 34279, {\it Automated Texture Generation for E-commerce}) and Cyanapse Limited (Brighton, United Kingdom) in collaboration with ZEG.ai Ltd. (London, United Kingdom).
\end{ack}

{\small
\bibliographystyle{plainnat}
\bibliography{egbib}
}


\end{document}


\maketitle


\newpage

\begin{table}
\small
\begin{center}
\begin{tabular}{l c c c}
  & Sample No. & Predicted DISTS & Measured DISTS \\
 \\
\hline
\\
Best 5  & 1   & 0.2523 & 0.2750\\
 & 2 & 0.2547  & 0.2798\\
 & 3 & 0.2941 & 0.2966\\
 & 4 & 0.2960 & 0.2965\\
 & 5 & 0.2969 & 0.3132\\
 \\
 \hline
\\
Average 5  & 1  & 0.3697 & 0.3623\\
 & 2  & 0.3699 & 0.3921\\
 & 3  & 0.3700 & 0.3529\\
 & 4  & 0.3700 & 0.3965\\
 & 5  & 0.3719 & 0.3510\\
 \\
 \hline
 \\
Worst 5 &  1  & 0.4373  & 0.4427\\
 & 2  & 0.4362 & 0.4535\\
 & 3  & 0.4340 & 0.5060\\
 & 4  & 0.4210 & 0.4483\\
 & 5  & 0.4190 & 0.4764\\
 \\
\hline
\end{tabular}
\bigskip
\caption{Five best (lowest), average,  and worst (highest) predicted DISTS scores for the samples from the day to night translation test dataset. Corresponding example images are shown in Figures S1, S2 and S3, respectively.}
\label{table:benchmark2}
\end{center}
\end{table}



\begin{figure}
\begin{center}
\includegraphics[width=0.9\linewidth]{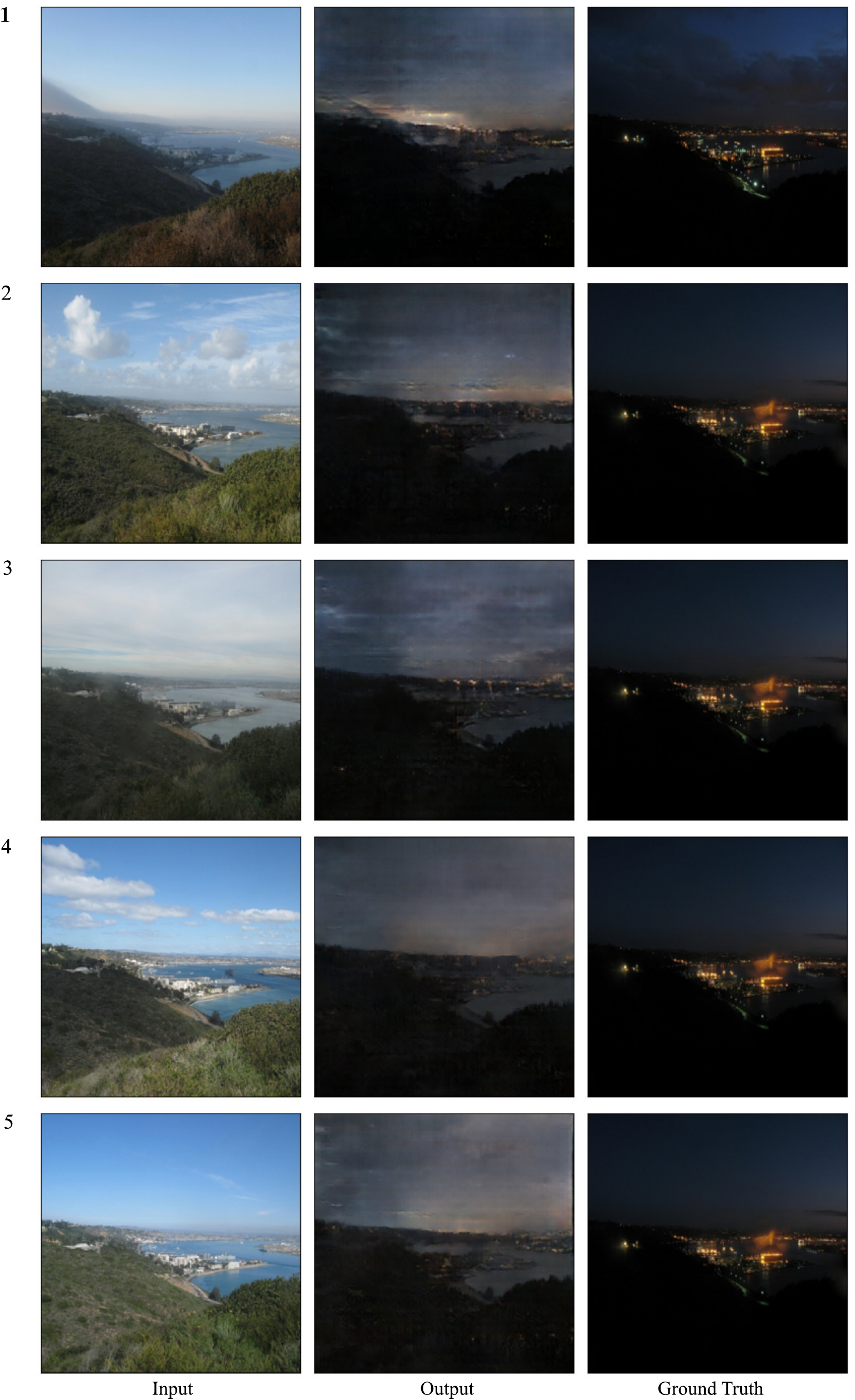}
\end{center}
\caption{Samples with the best (lowest) DISTS scores.}
\label{fig:sm_best}
\label{fig:onecol1}
\end{figure}

\begin{figure}
\begin{center}
\includegraphics[width=0.9\linewidth]{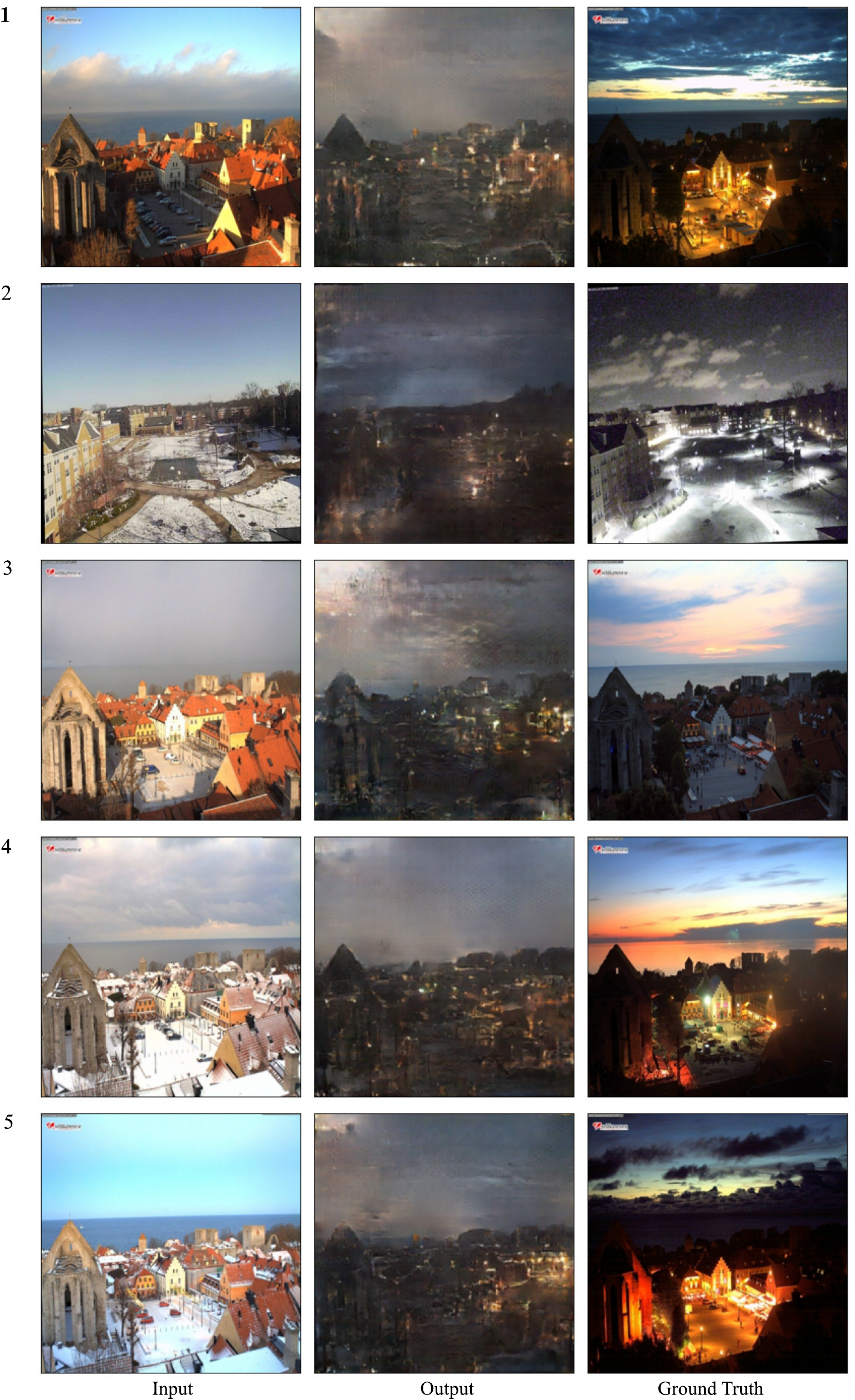}
\end{center}
\caption{Samples with average DISTS scores.}
\label{fig:sm_mid}
\label{fig:onecol1}
\end{figure}

\begin{figure}
\begin{center}
\includegraphics[width=0.9\linewidth]{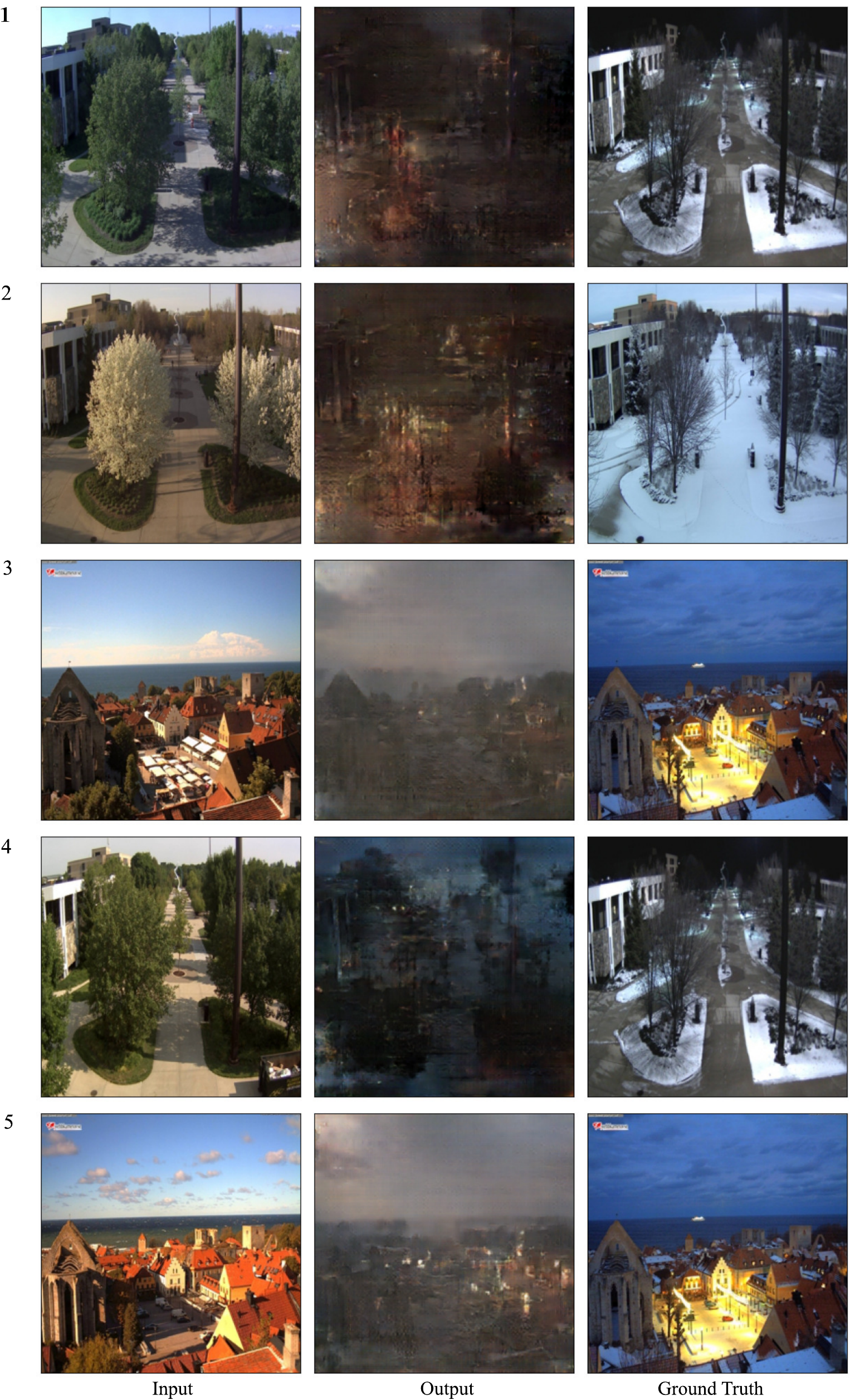}
\end{center}
\caption{Samples with the worst (highest) DISTS scores.}
\label{fig:sm_worst}
\label{fig:onecol1}
\end{figure}